\documentclass[letterpaper, 10 pt, conference]{template/ieeeconf}  

\IEEEoverridecommandlockouts                              

\usepackage{amsmath} 
\usepackage{amssymb}  

\usepackage{graphicx}
\usepackage[caption=false]{subfig}
\usepackage{hyphenat}
\usepackage[table]{xcolor}
\usepackage[font=small,belowskip=0pt, aboveskip=6pt]{caption}
\usepackage{multirow}
\usepackage{booktabs,amsfonts,dcolumn}
\usepackage{tabularx}
\usepackage{makecell}
\usepackage{hyperref}
\usepackage{comment}
\usepackage[style=ieee,mincitenames=1,maxcitenames=2,maxbibnames=10,doi=false,isbn=false,url=false,eprint=false]{biblatex}

\usepackage{xspace}

\long\def\invis#1{}

\newcommand\fig[1]{Fig.~\ref{#1}}

\newcommand\figref[1]{Fig. #1}

\makeatletter
\DeclareRobustCommand\onedot{\futurelet\@let@token\@onedot}
\def\@onedot{\ifx\@let@token.\else.\null\fi\xspace}

\def\etal{\emph{et al}\onedot}
\makeatother

\usepackage[group-separator={,}]{siunitx}

\title{\LARGE \bf
Real-Time Dense 3D Mapping of Underwater Environments 
}

\author{Weihan Wang$^{a*}$, Bharat Joshi$^{b*}$, Nathaniel Burgdorfer$^{a}$,  Konstantinos Batsos$^{c}$,  \\
Alberto Quattrini Li$^d$, Philippos Mordohai$^a$, Ioannis Rekleitis$^b$%
\thanks{$^*$ The first two authors have contributed equally to the paper.}%
\thanks{$^a$Stevens Institute of Technology, Hoboken, NJ, USA, 07030, {\tt\small \{wwang103,nburgdor,pmordoha\}@stevens.edu}}%
\thanks{$^b$University of South Carolina, Columbia, SC, USA, 29208, {\tt\small bjshi@email.sc.edu,  yiannisr@cse.sc.edu.}}%
\thanks{$^c$latitude AI, Palo Alto, CA, USA, 94304 {\tt\small kbatsos@stevens.edu}}%
\thanks{$^d$Dartmouth College, Hanover, NH, USA, 03755 {\tt\small alberto.quattrini.li@dartmouth.edu}}%
\thanks{
This research has been supported in part by the National Science Foundation under grants 1943205, 1919647, 2024741, 2024541 and 2024653. The authors would also like to acknowledge the help of the Woodville Karst Plain Project (WKPP) and El Centro Investigador del Sistema Acuífero de Quintana Roo A.C. (CINDAQ) in collecting data, providing access to challenging underwater caves, and mentoring us in underwater cave exploration. K. Batsos's contributions were made while at Stevens. 
} %
}

\begin{document}

\begin{minipage}{0.90\textwidth}\ \\[12pt]
\vspace{3in}
\begin{center}
     This paper has been accepted for publication in \textit{IEEE Conference on Robotics and Automation 2023}.  
\end{center}
  \vspace{1in}
  ©2023 IEEE. Personal use of this material is permitted. Permission from IEEE must be obtained for all other uses, in any current or future media, including reprinting/republishing this material for advertising or promotional purposes, creating new collective works, for resale or redistribution to servers or lists, or reuse of any copyrighted component of this work in other works.
\end{minipage}

\newpage

\maketitle

\begin{abstract}
This paper addresses real-time dense 3D reconstruction for a resource-constrained Autonomous Underwater Vehicle (AUV). Underwater vision-guided operations are among the most challenging as they combine 3D motion in the presence of external forces, limited visibility, and absence of global positioning. Obstacle avoidance and effective path planning require online dense reconstructions of the environment. Autonomous operation is central to environmental monitoring, marine archaeology, resource utilization, and underwater cave exploration. To address this problem, we propose to use SVIn2, a robust VIO method, together with a real-time 3D reconstruction pipeline. We provide extensive evaluation on four challenging underwater datasets. Our pipeline produces comparable reconstruction with that of COLMAP, the state-of-the-art offline 3D reconstruction method, at high frame rates on a single CPU.

\end{abstract}

\section{INTRODUCTION}
\label{sec:intro}

Mapping underwater environments is an important and challenging endeavor. 
Monitoring the coral reefs~\cite{williams2004}, exploring underwater caves~\cite{kernagis2008dive} and recording the shape of \textit{Cenotes}~\cite{gary20083d} have tremendous significance in our understanding and awareness of the environment.
Underwater mapping is also crucial for marine archaeology~\cite{rissolo2015novel}, infrastructure maintenance, and during search and rescue missions. Automating mapping with Autonomous Underwater Vehicles (AUVs) reduces risks to divers, enables longer operations times and increases the frequency of mapping/exploration missions.

Unfortunately, as demonstrated in recent work on comparing numerous open\hyp source visual and visual/inertial state estimation packages~\cite{QuattriniLiISERVO2016,JoshiIROS2019}, there are frequent failures underwater due to a variety of reasons. In contrast to above-water scenarios, GPS based localization is impossible. In addition to the traditional difficulties of vision based localization, the underwater environment is prone to rapid changes in lighting conditions, limited visibility, and loss of contrast and color information with depth~\cite{SkaffBMVC2008,roznere2019color}.

In this paper, we focus on real-time, scalable, detailed 3D mapping. These goals must be accomplished on a computational platform suitable for deployment on an AUV. Our software requires only a CPU and follows a pipeline architecture that incurs an almost constant computational load by processing fixed-length segments of the data at a time. The proposed approach includes: 
robust real-time camera pose estimation using SVIn2 \cite{RahmanIJRR2022} which fuses information from the cameras and IMU;
two-stage depth map estimation based on multi-threaded CPU-based stereo matching followed by visibility-based depth map fusion; and colored point cloud generation.

We conducted a thorough evaluation comparing our method to COLMAP \cite{colmap,schonberger16}, which is the state-of-the-art open-source 3D reconstruction framework. Our evaluation considers run-time, depth map estimation, and  dense reconstruction on four challenging underwater datasets. The main contributions of this paper lie on the integration of a state estimation and 3D reconstruction pipelines and the analysis of the feasibility of real-time dense reconstruction onboard.

\begin{figure}[t]
\footnotesize
    \centering
    \begin{tabular}{cc}
        \includegraphics[width=0.22\textwidth]{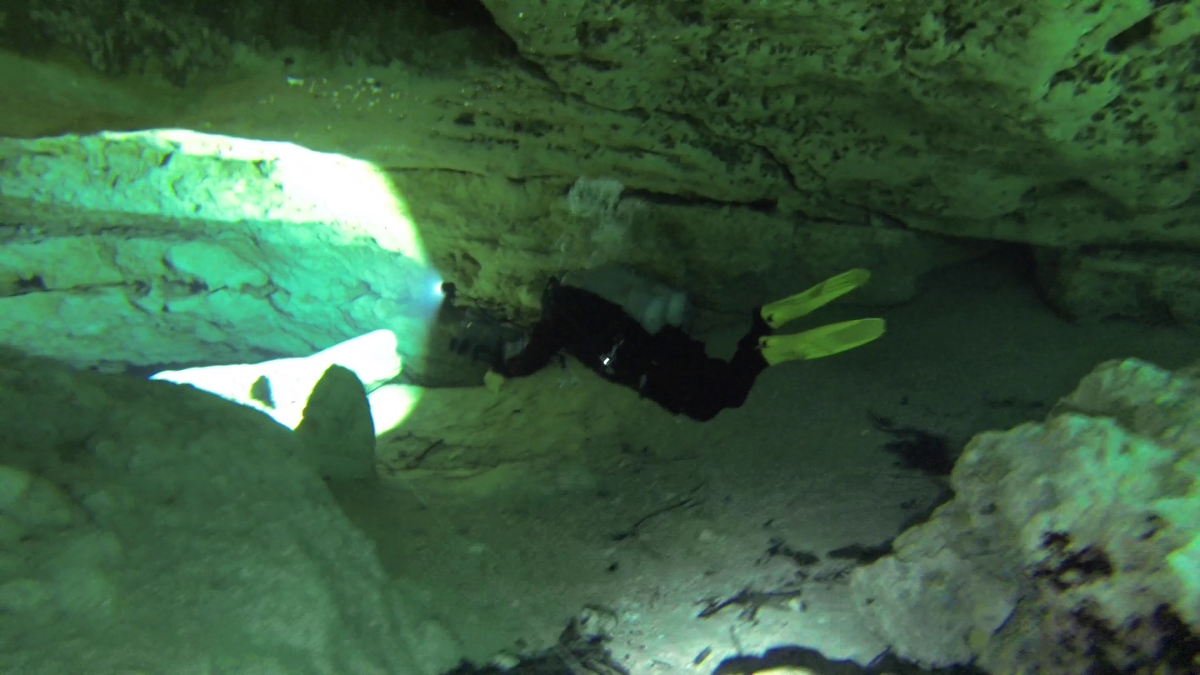}&
        \includegraphics[width=0.22\textwidth]{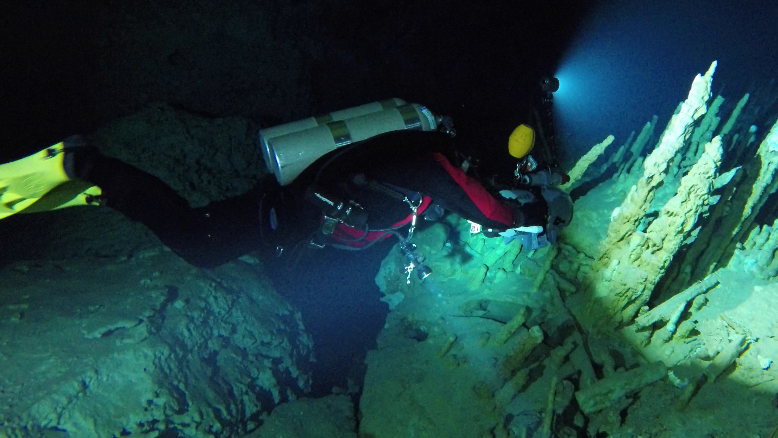}\\
        (a) Ginnie Ballroom, FL, USA & (b) Cenote, QR, Mexico \\
        \includegraphics[width=0.22\textwidth]{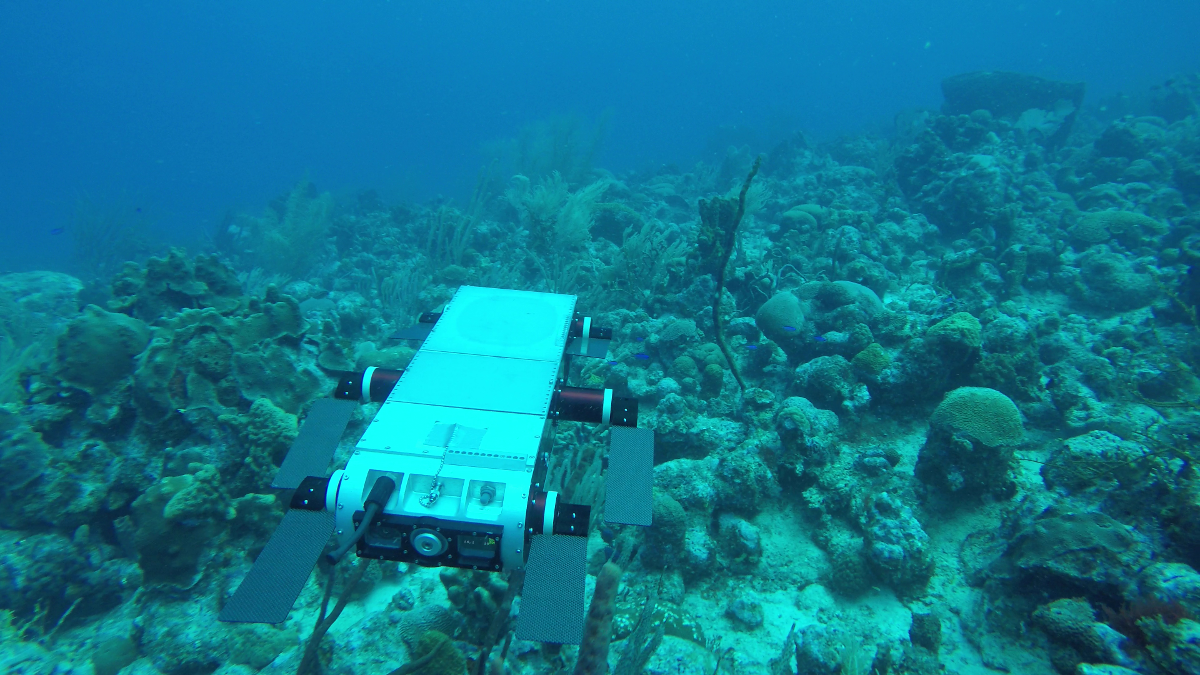}&
        \includegraphics[width=0.22\textwidth]{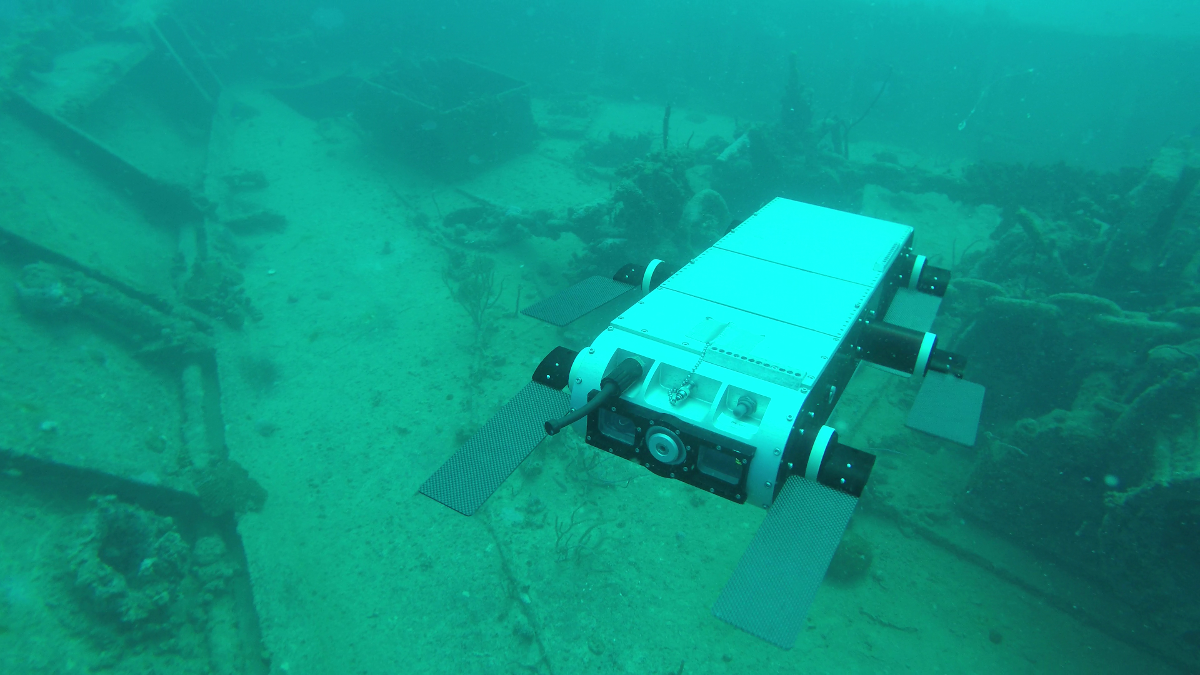} \\
        (c) Coral Reef, Barbados & (d) Stavronikita, Barbados
    \end{tabular}
    \caption{Datasets used in our experiments: (a) and (b) collected using a custom sensor suite. (c) and (d) collected using the Aqua2 AUV.}
    \label{fig:vehicles}
\end{figure}

\section{Related Work}
\label{sec:related}




The Structure-from-Motion (SfM) and Simultaneous Localization and Mapping (SLAM) literature is vast. Here, we focus on approaches tailored for underwater deployment.
State estimation underwater is challenging due to color saturation, floating particulates, and limited visibility \cite{JoshiIROS2019}. Vargas \etal \cite{vargas_robust_sensing} proposed robust visual SLAM underwater leveraging  acoustic, inertial and altimeter/depth sensors in addition to cameras. Tightly coupled fusion of visual, inertial, and pressure sensors using forward and backward IMU preintegration is discussed in \cite{hu_pressure_fusion}. We use the approach of Rahman \etal \cite{RahmanIJRR2022} to obtain robust camera pose estimates by fusing visual and inertial
information in real time.
%
Beall \etal. \cite{Beall} demonstrate accurate sparse 3D reconstruction of underwater structures
from stereo videos. Joshi \etal \cite{JoshiICRA2022} augment a visual SLAM algorithm 
so that, after loop closures, the map is deformed to preserve the relative pose between each point and its attached keyframes. 

Among the few authors that tackle dense underwater stereo, Queiroz-Neto et al. \cite{underwater_stereo}  model light propagation to overcome poor contrast and illumination.
Relevant to our method are general-purpose dense 3D reconstruction algorithms 
\cite{pollefeys08,stuhmer10,newcombe11,wendel12,pradeep2013monofusion,pizzoli14,zienkiewicz2016,schops17,teixeira2017,mordohai20} operating in online mode on the frames of one or more video streams. They can achieve high throughput, in many cases by leveraging powerful GPUs. They have been evaluated qualitatively since appropriate benchmarks with ground truth, other than KITTI \cite{Geiger2013IJRR}, 
are not available. 
Recently, learning-based approaches \cite{liu2019neural,xie2020video,duzceker2021deepvideomvs,long2021multi,min2021voldor+,sun2021neuralrecon} have shown promising results at high frame rates. Considering the lack of ground truth data from relevant domains, training these algorithms in supervised mode is practically impossible.

\begin{figure}[!ht]
    \begin{center}
    \includegraphics[width=\columnwidth]{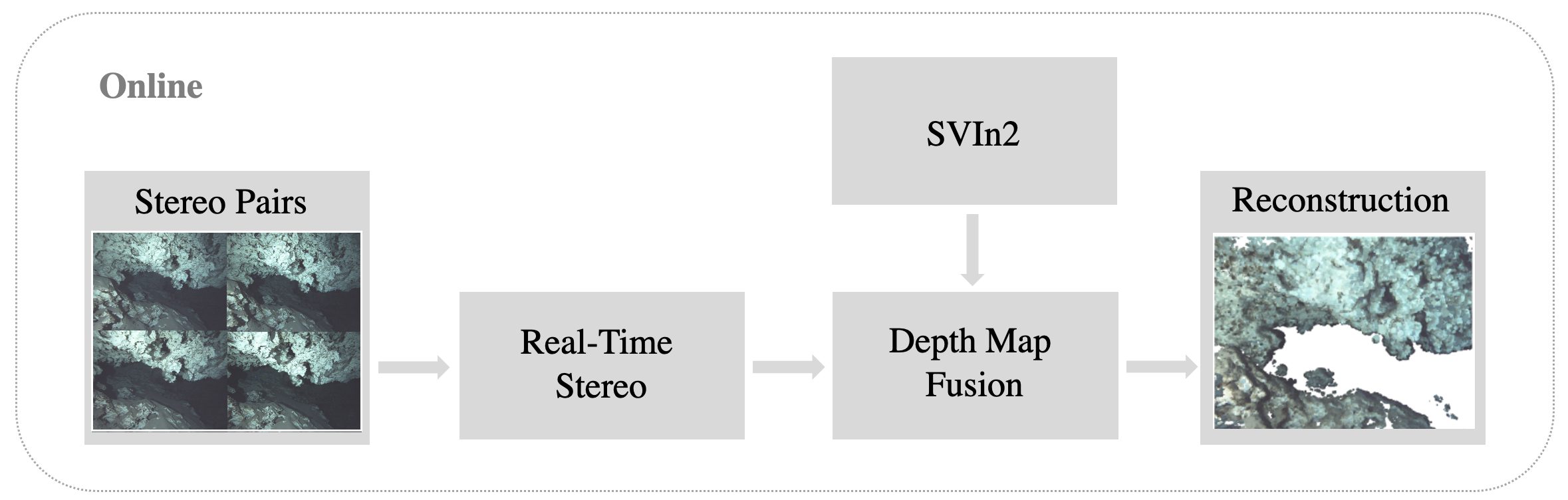}
    \end{center}
    \caption{A diagram of the components of the proposed pipeline. Given a pair of stereo images, we can estimate a depth map as well as the pose of the camera, using SVIn2, in parallel. Once both become available, depth maps can be fused to generate the final point cloud. }
    \label{fig:diagram}
\end{figure}

\section{Proposed Approach}\label{sec:method}

Our approach to 3D reconstruction relies on two parallel components: (1) a multi-sensor SLAM system, SVIn2 \cite{RahmanIJRR2022}, and (2) a real-time dense 3D mapping system, as shown in \fig{fig:diagram}. In this paper, we focus on the latter, as well as a comprehensive comparison of our online system with COLMAP \cite{colmap,schonberger16}. Our approach requires a calibrated stereo camera rig and an IMU to provide the necessary inputs. Moreover, SVIn2 can also take inputs from sonar and pressure sensors.

\invis{
Our approach requires a calibrated stereo camera rig, an IMU, and a pressure sensor to provide the necessary inputs.
}
%
%



\subsection{Pose Estimation}\label{sec:pose}
Robot pose estimation relies on our previous work, SVIn2~\cite{RahmanIROS2019b,RahmanIJRR2022}, a tightly-coupled keyframe-based SLAM system that fuses data from the cameras and IMU. 
We have demonstrated that SVIn2 performs well in underwater environments by explicitly addressing drift, loss of localization and poor illumination via robust initialization, loop-closing, and relocalization capabilities. 

\subsection{Depth Map Estimation}\label{sec:stereo}


The  stereo matching module estimates depth for every pixel of the left image of a stereo pair of images. The cameras are placed with their image planes approximately parallel facilitating rectification in software via a pair of homographies \cite{Hartley2004} estimated using a calibration checkerboard. This configuration allows the use of very fast algorithms for dense correspondence estimation that operate on horizontal epipolar lines. Metric depth can be obtained for each stereo frame, due to the known camera calibration parameters, regardless of the accuracy in camera pose estimation.

We are able to process stereo pairs at high throughput leveraging multi-threaded CPU implementations (OpenMP) without relying on GPUs. We accomplish this by carefully designing every step of the stereo matching process.
Our implementation of matching cost computation is publicly available\footnote{\url{https://github.com/kbatsos/Real-Time-Stereo}}.


\noindent \textbf{Matching cost computation.} Stereo matching estimates the likelihood \cite{hirschmuller2007evaluation} for each possible \textit{disparity} that can be assigned to a given pixel of the reference image, typically the left. 
Disparity $d$ is defined as the difference between the horizontal coordinates of two potentially corresponding pixels in the same epipolar line (scanline) in the left and right image. Disparity  is inversely proportional to depth $Z$, which can be computed as:
\begin{equation}
    Z = \frac{bf}{d}
\end{equation}
\noindent where $b$ is the baseline (distance) between the camera centers and $f$ is the focal length.

To select the most likely disparity for each pixel in the left image, we assess the photoconsistency of that pixel with potentially matching pixels in the conjugate epipolar line in the right image. This is accomplished by computing a similarity or dissimilarity measure in matching windows centered around the pixels under consideration.
In this paper, we experimented with the Sum of Absolute Differences (SAD), which is a dissimilarity (cost).
The cost values for all pixels and disparities are accumulated in the \textit{cost volume} $V$, which is computed as follows with SAD:
\begin{equation}
    V(x_L,y,d) = \sum_{(u,v)\in W(x_L,y)} |I_L(u,v)-I_R(u-d,v)|
\end{equation}
\noindent where $I_L$ and $I_R$ are the two images and $W(x_L,y)$ is the matching window.

We accelerate these computations using several techniques including: \textit{integral images} to compute sums in rectangular sub-images in constant time \cite{veksler2003fast} 
regardless of the matching window size; careful design of the memory layout of all data; and memoization.
%

\noindent \textbf{Optimization.} 
The fastest way to obtain a disparity map from the cost volume is by selecting the disparity with the smallest cost for each pixel. 
To obtain higher accuracy the cost volume can be optimized by the widely-used Semi-Global Matching algorithm (SGM) \cite{hirschmuller08}.

SGM is used for extracting a disparity map that approximately optimizes a global cost function defined over 2D image neighborhood by combining multiple 1D cost minimization problems. Briefly, SGM favors constant disparity, imposes a small penalty for disparity differences equal to 1 between adjacent pixels along the minimization direction, and imposes a larger penalty for large discontinuities. This has the effect of allowing slanted surfaces and reducing the number of jumps in disparity. 
%
Here, we integrate the rSGM implementation of Spangenberg et al. \cite{spangenberg2014large} opting for the variant that considers only four 1D sub-problems to favor speed.
%
Disparity is converted to depth, which is then refined to sub-pixel precision by fitting a parabola in the vicinity of the minimum optimized cost \cite{scharstein02}.

\noindent \textbf{Confidence estimation.} Depth map fusion benefits from confidence values conveying which depths are more reliable. We attach a confidence to each depth after SGM using the PKRN measure \cite{hu12}, which is the ratio of the second smallest over the smallest cost for that pixel in the cost volume. PKRN is effective in discriminating reliable from unreliable depths and can be computed in negligible time during the final disparity selection step.

\subsection{Depth Map Fusion}\label{sec:fusion}

Depth maps estimated by the stereo matching module are reasonably accurate but contain noise due to lack of texture, occlusion, and motion blur. Under the assumption that the errors are not systematic and do not form hallucinated surfaces, 
we can improve the depth maps by fusing them. 
This approach takes as input overlapping depth and detects consensus among depth estimates and violations of visibility constraints as evidence for which depths are correct and which are likely to be outliers.


Similar to our previous work \cite{merrell07,hu2012least}, the input for computing a fused depth map for a given \textit{reference view} is a set of $N_f$ depth maps and the corresponding confidence maps in a sliding window of frames. The middle depth map is used as reference and all depth and confidence maps are rendered onto it.
%
We denote depths rendered on the reference view by $Z_j$ and depths in their original camera coordinate systems by $Z^o_j$. 
For each of the depth candidates $Z_j$ we accumulate \textit{support} and \textit{visibility violations}. 

\begin{figure}[t]
\vspace{0.075in}
    \begin{center}
    \includegraphics[width=.95\columnwidth]{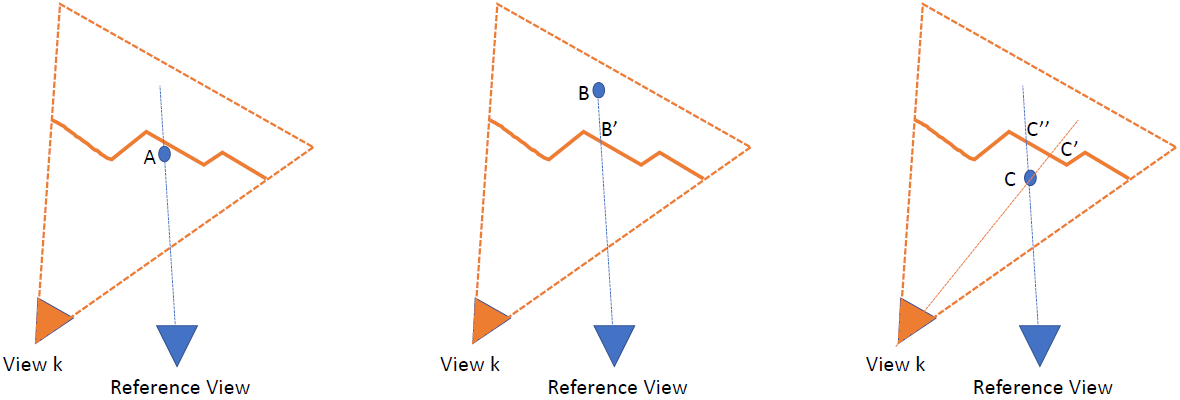}
    \end{center}
    \caption{Illustration of depth map fusion. Points A, B and C are depth candidates for pixels of the reference view, estimated directly or rendered to it from other views. The orange line marks the cross section of the surface estimated by view $k$. Left: point A is supported by the orange surface. Middle: point B is occluded by B', which is in front of B in the ray of the reference view. Right: point C violates the free space of C' on the ray of view $k$. (Note that there is no conflict between C and C''.)} 
    \label{fig:fusion}
\end{figure}

Support comes from other depth candidates for the same pixel that are within a small distance of $Z_j$.
$Z_j$ is then replaced by the weighted average of the supporting depths, with confidence values serving as weights. The confidence of the blended depth estimate $Z^s_j$ is set equal to the sum of the supporting confidences. See Fig.~\ref{fig:fusion}~(left).
\begin{align}
    Z^s_j &= \frac{s_{ij}Z_i}{s_{ij}}, \ \ \ s_{ij}=T(||Z_i-Z_j||<\epsilon) \label{eq:fuze_z}\\
    C^s_j &= \frac{s_{ij}C_i}{s_{ij}}, \ \ \ s_{ij}=T(||Z_i-Z_j||<\epsilon) \label{eq:fuze_conf}
\end{align}
\noindent where $s_{ij}$ is a boolean variable indicating whether $Z_i$ and $Z_j$ support each other and $T()$ is the indicator function which is 1 when its argument is true.

There are two types of violations of visibility constraints: occlusions and free space violations. An \textit{occlusion} occurs when $Z_j$ appears behind a rendered depth map from view $k$, $Z_k$,  on the ray of the reference view, as in Fig.~\ref{fig:fusion}~(middle), while a \textit{free space violation} occurs when $Z_j$ appears in front of an input depth map $Z^o_l$ on the ray of
view $l$, as in Fig.~\ref{fig:fusion}~(right). 
Detected violations do not result in updates to the fused depth, but the corresponding confidence is reduced by the confidence of the conflicting estimate, similarly to Eq. (\ref{eq:fuze_conf}).
We assign to each pixel the depth with the highest fused confidence, after adding support and subtracting conflicts. We then threshold confidence to reject outliers.

The fusion process is independent across pixels, and is thus parallelizable. Rendering depth candidates to the original depth maps to detect free space violations is the most expensive step.
Fusion operates in pipeline mode keeping a small number of recent depth maps in memory at a given time. At the next time step, the oldest depth map in the sliding window is dropped and it is replaced by the most recent one. As discussed in Section \ref{sec:results}, the sliding window we use is very short to keep latency low.

\section{Experimental Results} \label{sec:results}

In this section, we present experimental results on challenging underwater sequences. We evaluate both the sparse and dense components of the 3D reconstruction system and compare them to the corresponding aspects of COLMAP, which operates much slower in offline mode.

\subsection{Datasets}\label{sec:datasets}
The datasets used in this paper were collected using a custom made sensor suite~\cite{RahmanOceans2018}; see \fig{fig:vehicles}(a) and (b), and an Aqua2 robot~\cite{DudekIROS2005}; see \fig{fig:vehicles}(c) and (d). Both devices are equipped with two iDS USB3 uEye cameras as stereo pair, a MicroStrain 3DM-GX4-15 IMU, and a Bluerobotics Bar30 water pressure sensor. The stereo images are recorded at \SI{15}{\Hz}; inertial data at \SI{100}{\Hz}; and water pressure at \SI{1}{\Hz} using onboard Intel NUC. A video light is attached to the sensor suite unit to provide artificial illumination of the scene.  The Aqua2 AUV is a hexapod robot which utilizes the motion from six flippers, each actuated independently by an electric motor, to move in 3D.
\invis{The field of view of the cameras is $120^\circ$ (horizontal) and $90^\circ$ (vertical) tilted downward by $40^\circ$ from the horizontal plane.  The Aqua2 AUV is a hexapod robot, approximately $65{\rm cm} \times 45{\rm cm} \times 13{\rm cm}$ in size and $10{\rm kg}$ in weight. Underwater, Aqua2 utilizes the motion from six flippers, each actuated independently by an electric motor, to move in 3D.  Aqua2 has 6DoF, of which five are controllable, two directions of translation (forward/backward and upward/downward), along with roll, pitch and yaw.}

\begin{figure}[t]
\vspace{0.07in}
\footnotesize
    \centering
    \begin{tabular}{cc}
        \includegraphics[width=0.2\textwidth]{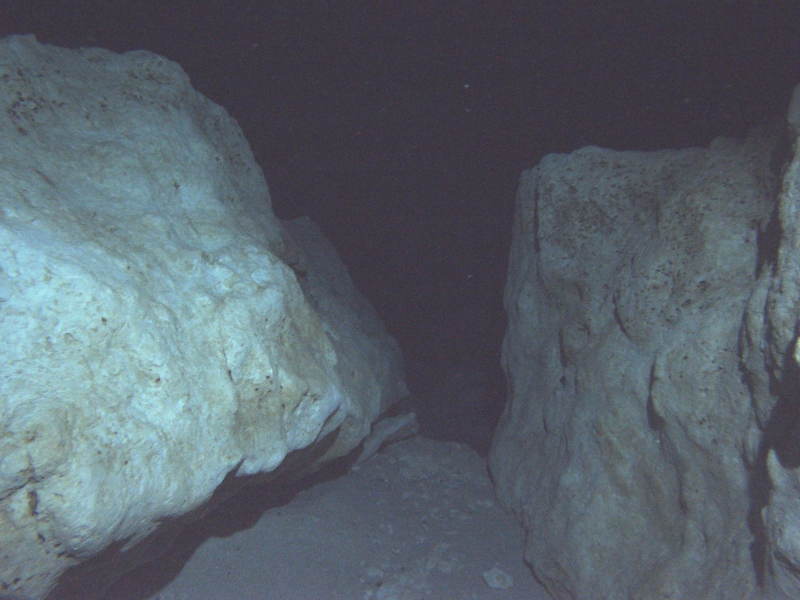}&
        \includegraphics[width=0.2\textwidth]{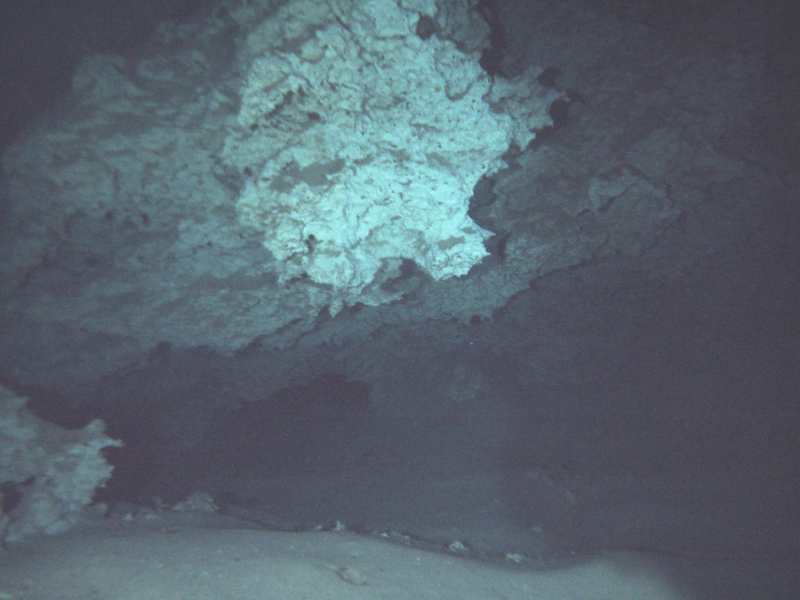}\\
        (a) Ginnie Ballroom & (b) Cenote \\
        \includegraphics[width=0.2\textwidth]{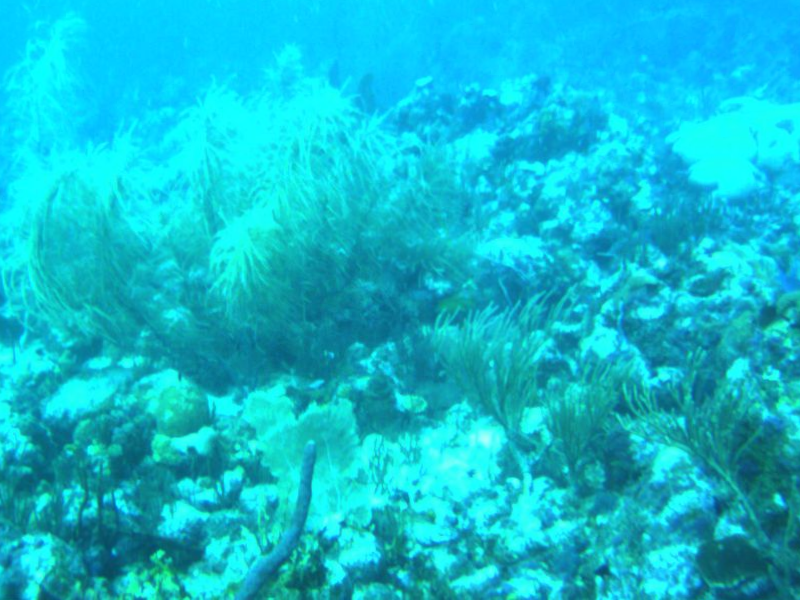}&
        \includegraphics[width=0.2\textwidth]{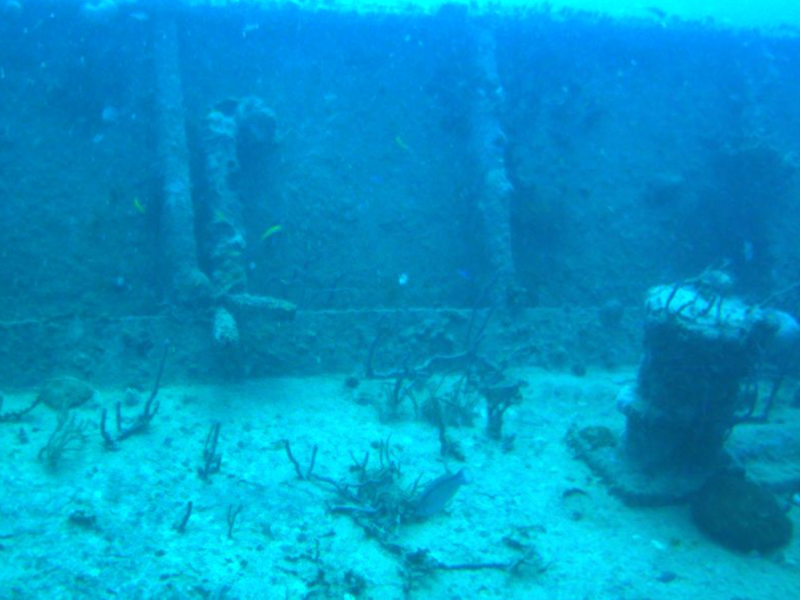} \\
        (c) Coral Reef & (d) Stavronikita
    \end{tabular}
    \caption{Sample images from the underwater datasets.}
    \label{fig:sampleImg}
\end{figure}

We perform experiments on four datasets:
\begin{itemize}
    \item \textbf{Ginnie Ballroom, Gennie Springs, FL, USA}
    \item \textbf{Cenote, QR, Mexico}
    \item \textbf{Coral Reef, Barbados}
    \item \textbf{Stavronikita Shipwreck, Barbados}
\end{itemize}
These underwater datasets present substantial challenges for 3D reconstruction. The \textit{Ginnie Ballroom} and \textit{Cenote} datasets are collected using a custom sensor suite \cite{RahmanOceans2018} operated by a diver, \fig{fig:sampleImg}(a) and (b); while the \textit{Coral Reef} and \textit{Stavronikita Shipwreck} datasets are collected using the Aqua2 AUV performing a lawnmower pattern over the scene, \fig{fig:sampleImg}(c) and (d). These datasets form a diverse set of underwater environments, including open, flat areas of the seafloor, dense and richly structured shipwrecks, and enclosed caverns with relatively uniform surfaces. In the \textit{Coral Reef} and \textit{Stavronikita Shipwreck} datasets, we can rely on natural light to illuminate the scene, but for the \textit{Ginnie Ballroom} and \textit{Cenote} datasets, we must rely on artificial illumination from the sensor suite.

\subsection{COLMAP}\label{sec:colmap}

In this section, we briefly describe COLMAP \cite{colmap,schonberger16}, a state-of-the-art, open-source Structure-from-Motion and dense 3D reconstruction software used as a baseline in our experiments. 
The sparse component of COLMAP takes as input an unordered image collection, extracts and matches features, builds the scene graph and performs bundle adjustment. 
The dense reconstruction component jointly estimates depth and surface normals using a PatchMatch Multi-View Stereo (MVS) algorithm with pixel-wise view selection, photometric and geometric priors. The 
depth maps are fused based on multi-view geometric consistency to produce a dense 3D reconstruction.

In our experiments, we pass the keyframes obtained from SVIn2 
to COLMAP to obtain bundle-adjusted camera poses, dense depth maps and point clouds. Unless otherwise noted, all dense reconstruction experiments are performed using these camera poses as input.



\begin{figure}[!ht]
\vspace{0.05in}
\footnotesize
    \begin{center}
    \resizebox{\columnwidth}{!}{
    \begin{tabular}{ccc}
        \includegraphics[width=0.52\columnwidth, trim={0.2in 0 0.05in 0},clip]{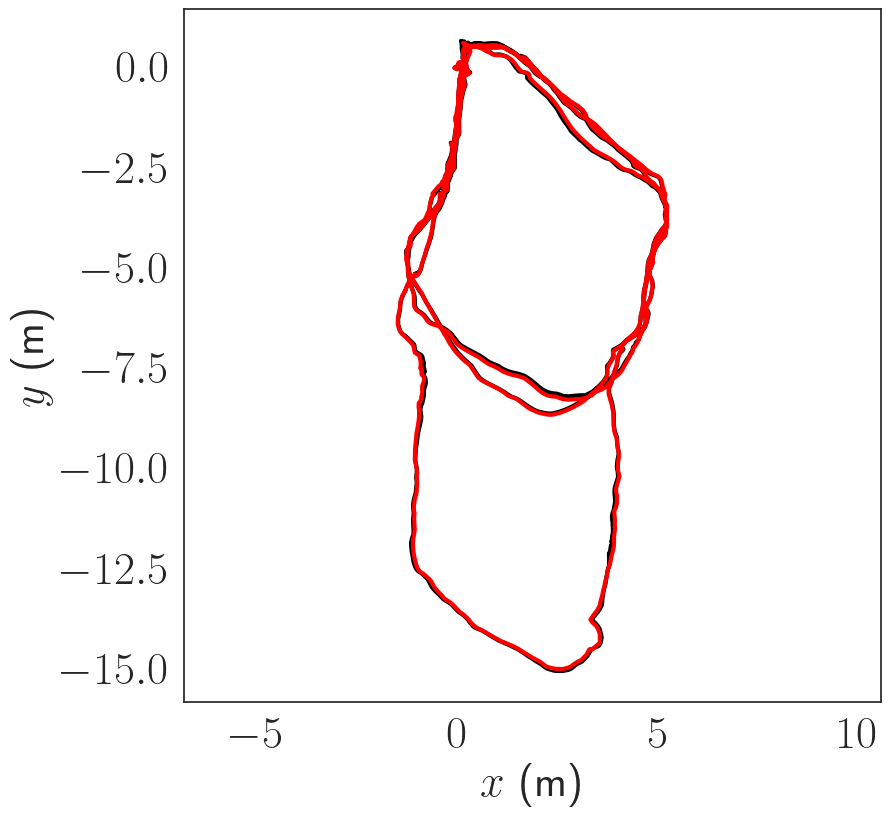}&
        \includegraphics[width=0.5\columnwidth, trim={0.2in 0 0.05in 0},clip]{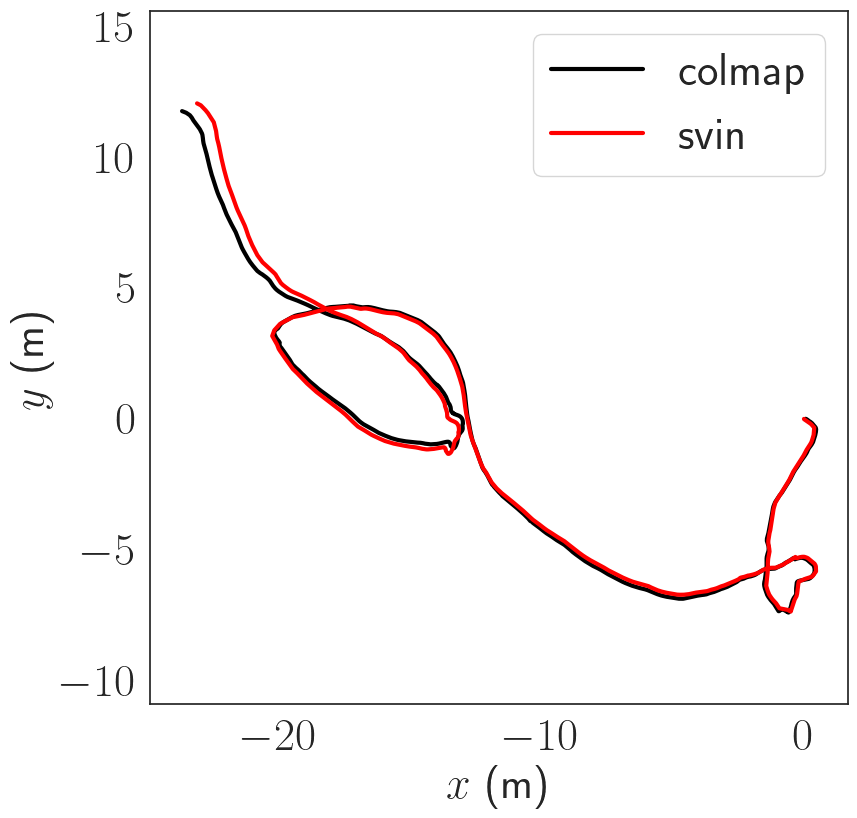}&
        \includegraphics[width=0.5\columnwidth, trim={0.2in 0 0.05in 0},clip]{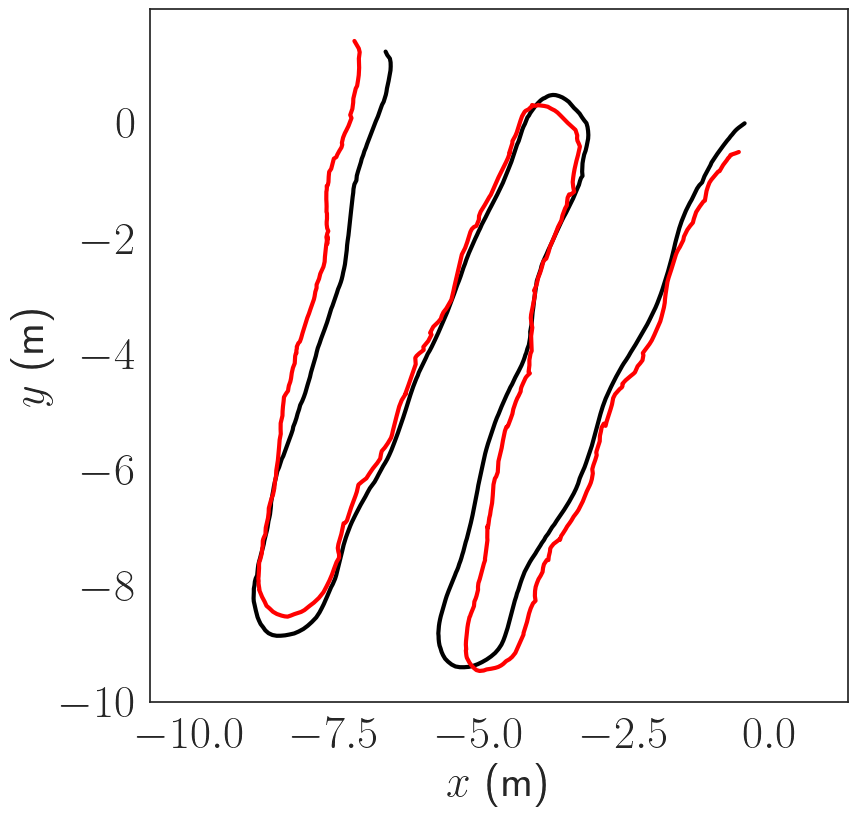} \\
    \end{tabular}
    }
    \end{center}
    \hspace{0.2in}(a) Ginnie Ballroom \hspace{0.3in}(b) Cenote  
    \hspace{0.45in}(c) Coral Reef

    \caption{Camera trajectories, estimated by  SVIn2~\cite{RahmanIROS2019a} and COLMAP~\cite{colmap} after \textit{sim3} alignment.}
    \label{fig:traj}
\end{figure}

\subsection{Camera Pose Estimation Results}\label{sec:tracking}

We compare two drastically different approaches for estimating the trajectory: COLMAP which operates offline and performs global bundle adjustment and SVIn2 which runs online performing SLAM. Due to the known baseline of the stereo camera and the inertial data, SVIn2 produces trajectories with correct scale. However, these trajectories may suffer from drift, especially when loop closure opportunities are unavailable. Due to global bundle adjustment, COLMAP trajectories are more accurate, but scale may drift during optimization. During post-processing, the scale discrepancy is corrected by scaling the camera poses from COLMAP using the known stereo baseline.

To enable comparisons, we provide the keyframes selected by SVIn2 as input to COLMAP  and align SVIn2 trajectory with COLMAP using \textit{sim3} alignment \cite{umeyama_alignment}. We use the root mean square Absolute Trajectory Error (ATE) metric ~\cite{scaramuzza_trajectory_evaluation} to compare the trajectories. Note that in the absence of ground truth, we can only measure the discrepancy between the COLMAP and SVIn2. \figref{\ref{fig:traj}} shows the trajectories estimated by COLMAP and SVIn2 after \textit{sim3} alignment, while 
Table~\ref{tab:rmse} shows the root mean square ATE in meters. The trajectories are in general consistent up to a few cm in terms of ATE. 

We were unable to obtain a complete trajectory on the Stavronikita dataset using SVIn2 due to segments in which the AUV maneuvered over the side of the wreck, thus facing open water, causing SVIn2 to lose track. COLMAP, on the other hand, attempts to match features over all images and registers a lot of the frames bridging gaps. We only use the COLMAP trajectory for Stavronikita in the remainder.

\begin{table}[!b]
    \centering
    \caption{Comparison of the SVIn2 and COLMAP trajectories based on root mean square ATE in meters.}
    \label{tab:rmse}
    \begin{tabular}{@{}lcc}
      Dataset &  length[m] & rmse \\
     \toprule
     Ginnie Ballroom & 98.73 & 0.07 \\
     Cenote & 67.24 & 0.19 \\
     Coral Reef & 45.52 & 0.39  \\
     Stavronikita & 106.30 & N/A \\
    \bottomrule
    \end{tabular}
\end{table}

\subsection{Dense Reconstruction}\label{sec:recon_results}


Stereo matching is performed on 800$\times$600 images with 100 disparity levels using the Sum of Absolute Differences (SAD) as the matching function in 3$\times$3 windows. We generate a depth map for every frame after SGM optimization and sub-pixel fitting and fuse three depth maps using the middle frame as reference. 
We set the support radius during depth map fusion,  $\epsilon$, to 0.04 and the threshold on fused confidence for outlier rejection, $C_{\textrm{thres}}$, to 0.5. 


\noindent \textbf{Evaluation}. In the absence of ground truth and of any practical system for acquiring ground truth underwater, we evaluate our online reconstruction pipeline by comparing the output depth maps and 3D point cloud with those  generated offline by COLMAP. It should be noted that the latter are not perfect, but benefit from global optimization.


\begin{figure}
\vspace{0.08in}
\footnotesize
    \centering
    \begin{tabular}{cc}
        \includegraphics[width=0.4\columnwidth]{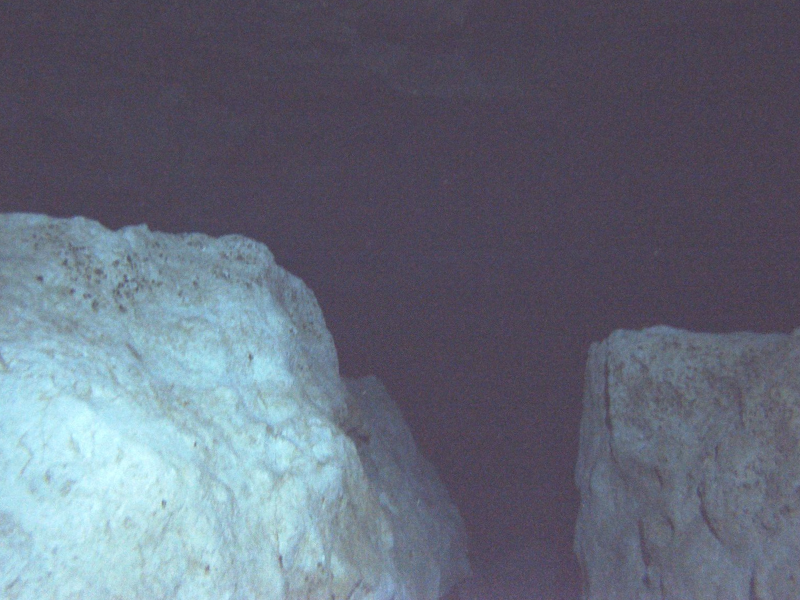}&
        \includegraphics[width=0.4\columnwidth]{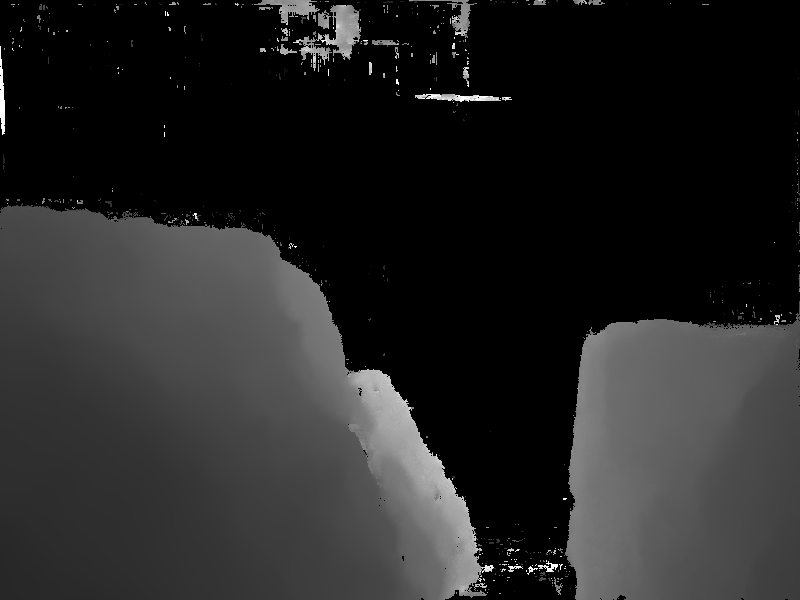}\\
        Image & COLMAP Depth \\
        \includegraphics[width=0.4\columnwidth]{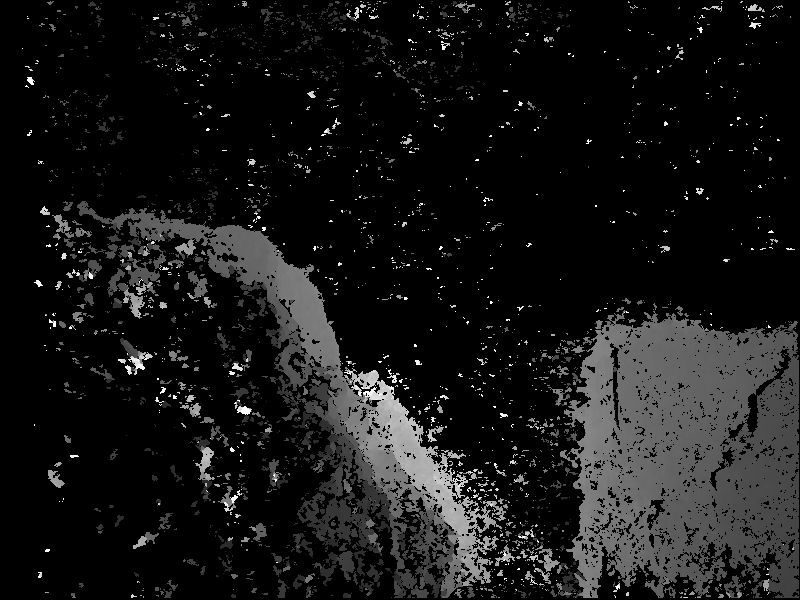}&
        \includegraphics[width=0.4\columnwidth]{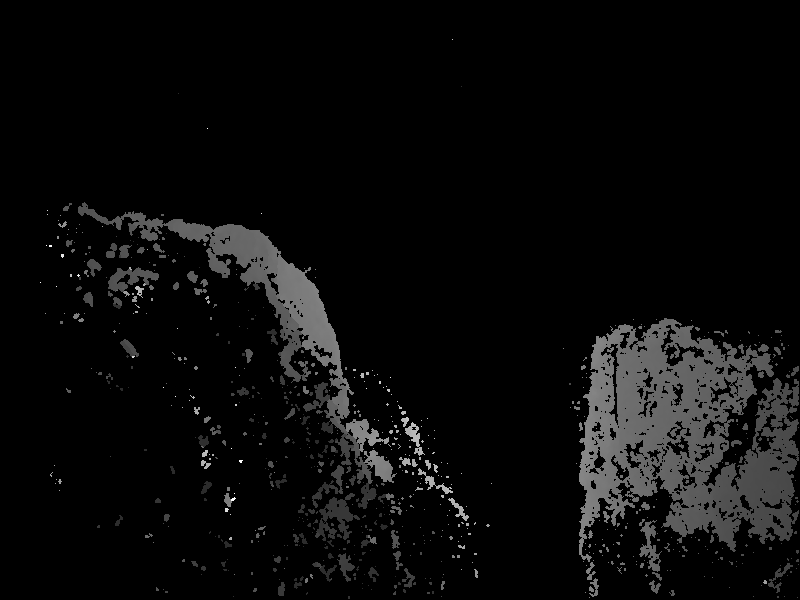}\\
        Pipeline Raw Depth & Pipeline Fused Depth
    \end{tabular}
    \caption{\textbf{Ginnie Ballroom dataset}. Example depth maps from COLMAP and Pipeline.}
    \label{fig:ginnie_depth}
\end{figure}

\noindent \textbf{Run-time comparison}. 
In the four datasets (Ginnie Ballroom, Cenote, Coral Reef, Stavronikita) that we consider, there are 13671, 3519, 3207, 8824 stereo pairs and 1519, 1401, 631 and 897 keyframes selected by SVIn2, respectively. We ran the pipeline on an Intel i7-10700K desktop with 32GB memory, while COLMAP is run on desktop with Intel i9-12900K CPU, 32GB memory, and an NVIDIA GeForce RTX 2080Ti GPU. Comparison of the run-times on the keyframes between COLMAP and our pipeline is listed in Table \ref{tab:run_times}.  
The pipeline outperforms COLMAP significantly with respect to speed; 
it achieves a throughput ranging between 2.8 and 10.2 fps  on the four datasets, whereas COLMAP runs at 0.05 - 0.3 fps, when considering all input frames (not just keyframes), which are a true measure of the length of the input videos.


\begin{table}[!bh]
\vspace{0.08in}
    \centering
    \caption{Depth map evaluation: Comparison of COLMAP-generated depth maps with \textit{raw} and \textit{fused} depth maps from the pipeline.  MoM (Median-of-Medians) is the median of the median per-depth map absolute errors. MAE is the per-pixel mean absolute depth error. All errors are in meters.
    }
    \label{tab:depth_map_results}
    \scalebox{0.75}{
    \begin{tabular}{l|cc|cc}
    \specialrule{.8pt}{0pt}{0pt}
    \multirow{2}{*}{\textbf{Dataset}} 
    &\multicolumn{2}{c|}{\textbf{Raw Depth Maps vs COLMAP}} &\multicolumn{2}{c}{\textbf{Fused Depth Maps vs COLMAP}}\\
    &MoM (m) & MAE (m) &MoM (m) & MAE (m) \\
    \hline
     Ginnie Ballroom &0.127 &0.452 & 0.063  & 0.079 \\
     Cenote  &0.188 &0.469 & 0.105 & 0.134 \\
     Coral Reef &0.400 &0.626 & 0.327 & 0.402\\
     Stavronikita &0.457&0.684 & 0.336 & 0.399\\
    \specialrule{.8pt}{0pt}{2pt}
    \end{tabular}
    }
\end{table}

\noindent \textbf{2D Metrics}. 
Both COLMAP and the pipeline produce dense depth maps, which, however, contain some holes without depth estimates due to filtering. Avoiding to generate noise, especially since the same surface may be reconstructed from a different view, is desirable. Therefore, we only compare depth estimates that exist in both depth maps by recording the absolute depth error (AE). We then compute the mean (MAE) and median of these errors per depth map, as well as the MAE over an entire dataset and the median-of-medians (MoM) as an approximation of the overall median over valid depths.


\noindent \textbf{3D Metrics}. To compare the point cloud reconstruction from the pipeline with the offline reconstruction generated by COLMAP, we utilize 
Chamfer distance metrics between the two models. We refer to the pipeline point cloud as the \textit{source} and the COLMAP point cloud as the \textit{target}. \textit{Accuracy} is the mean Chamfer distance from every point in the source model to the closest point in the target model. \textit{Completeness} is the mean Chamfer distance from every point in the target model to the closest point in the source model. 
\textit{Precision} and \textit{Recall} are the percentage of points that have a Chamfer distance to the other set below a threshold;  \textit{Precision} is measured from \textit{source-to-target} and \textit{Recall} is measured from \textit{target-to-source}. For our evaluation, we set the threshold to 0.1m.

\begin{figure}
\vspace{0.08in}
\footnotesize
    \centering
    \begin{tabular}{cc}
        \includegraphics[width=0.4\columnwidth]{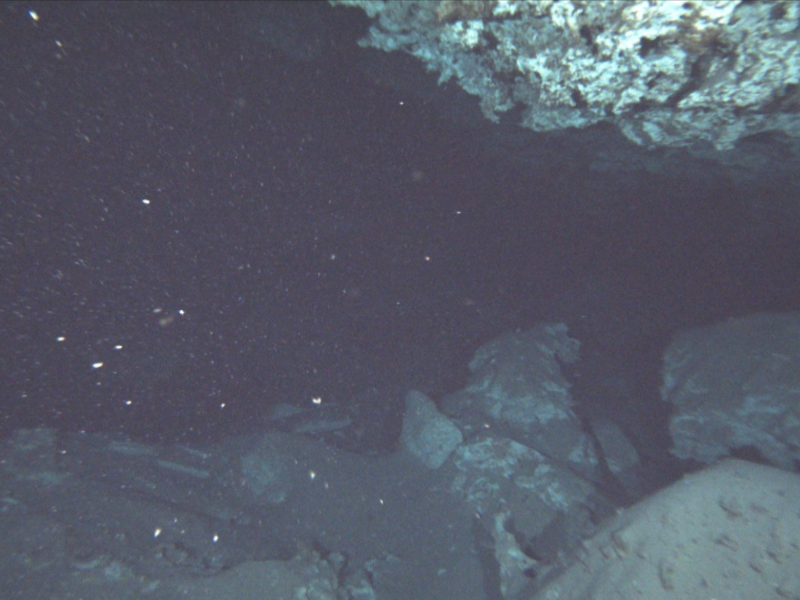}&
        \includegraphics[width=0.4\columnwidth]{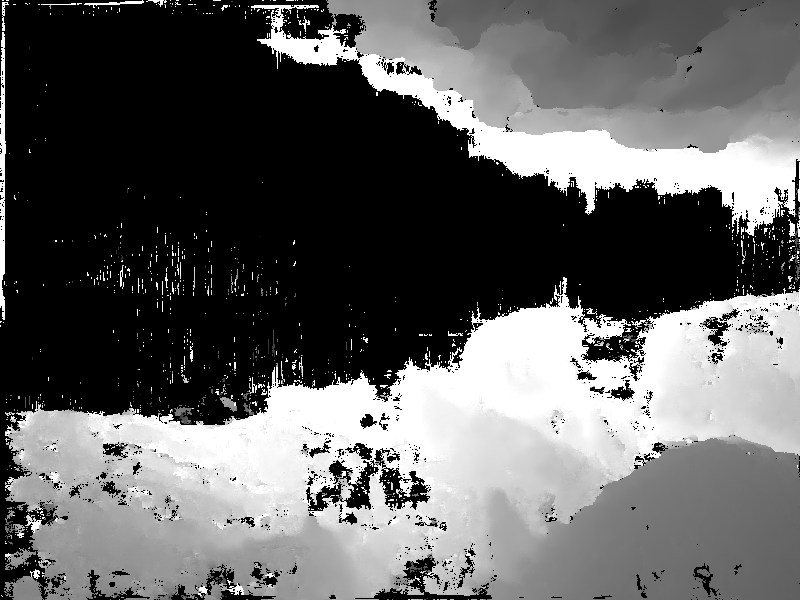}\\
        Image & COLMAP Depth \\
        \includegraphics[width=0.4\columnwidth]{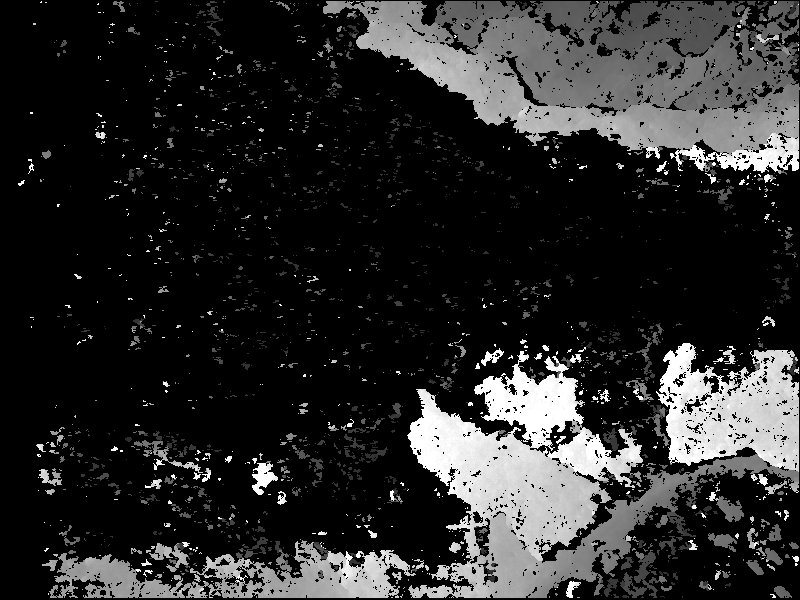}&
        \includegraphics[width=0.4\columnwidth]{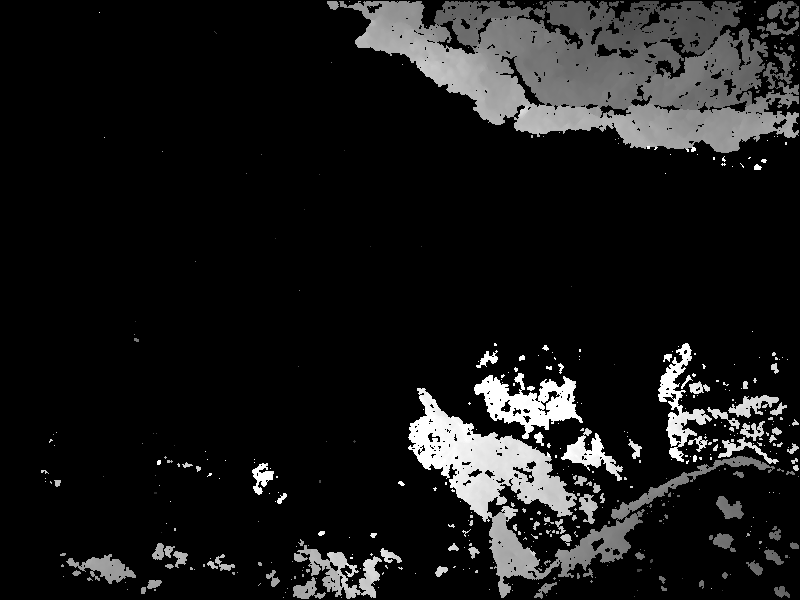}\\
        Pipeline Raw Depth & Pipeline Fused Depth
    \end{tabular}
    \caption{\textbf{Cenote dataset}. Example depth maps from COLMAP and Pipeline.}
    \label{fig:cenote_depth}
\end{figure}

\begin{table*}[!ht]
\vspace{0.2in}
    \caption{Run-time comparison between COLMAP and Pipeline.}
    \label{tab:run_times}
    \centering
    \scalebox{0.95}{
    \begin{tabular}{@{}l|c|rrr|rrrr}
    \specialrule{.8pt}{0pt}{0pt}
    \multirow{2}{*}{\textbf{Dataset}} & \multicolumn{1}{c|}{}& \multicolumn{3}{c|}{\textbf{COLMAP}} & \multicolumn{3}{c}{\hspace{1.5cm} \textbf{Pipeline}} \\& stereo pairs & vertices & MVS (min) & total time (min) & vertices & Stereo (ms/frame)&Fusion (ms/frame) & total time (min) \\
    \hline
     Ginnie Ballroom & \num{1519} & \num{8056361} & 607.61 & 858.10 & \num{54034304}& 398.88&374.93 & 22.36 \\
     Cenote & \num{1401} & \num{8909774} & 568.52 & 1250.53 &\num{91036005} &360.49&418.71 &20.92 \\
     Coral reef & 631 & \num{3833107} & 254.31 & 337.99 & \num{66733182} & 369.35 & 677.65 &12.44\\
     Stavronikita & 897 & \num{4552326} & 367.17 &482.57 & \num{29851344} & 323.57 & 553.37 & 14.95 \\ 
    \specialrule{.8pt}{0pt}{2pt}
    \end{tabular}
    }
\end{table*}

\begin{figure}[h]
\footnotesize
    \centering
       \includegraphics[width=0.82\columnwidth]{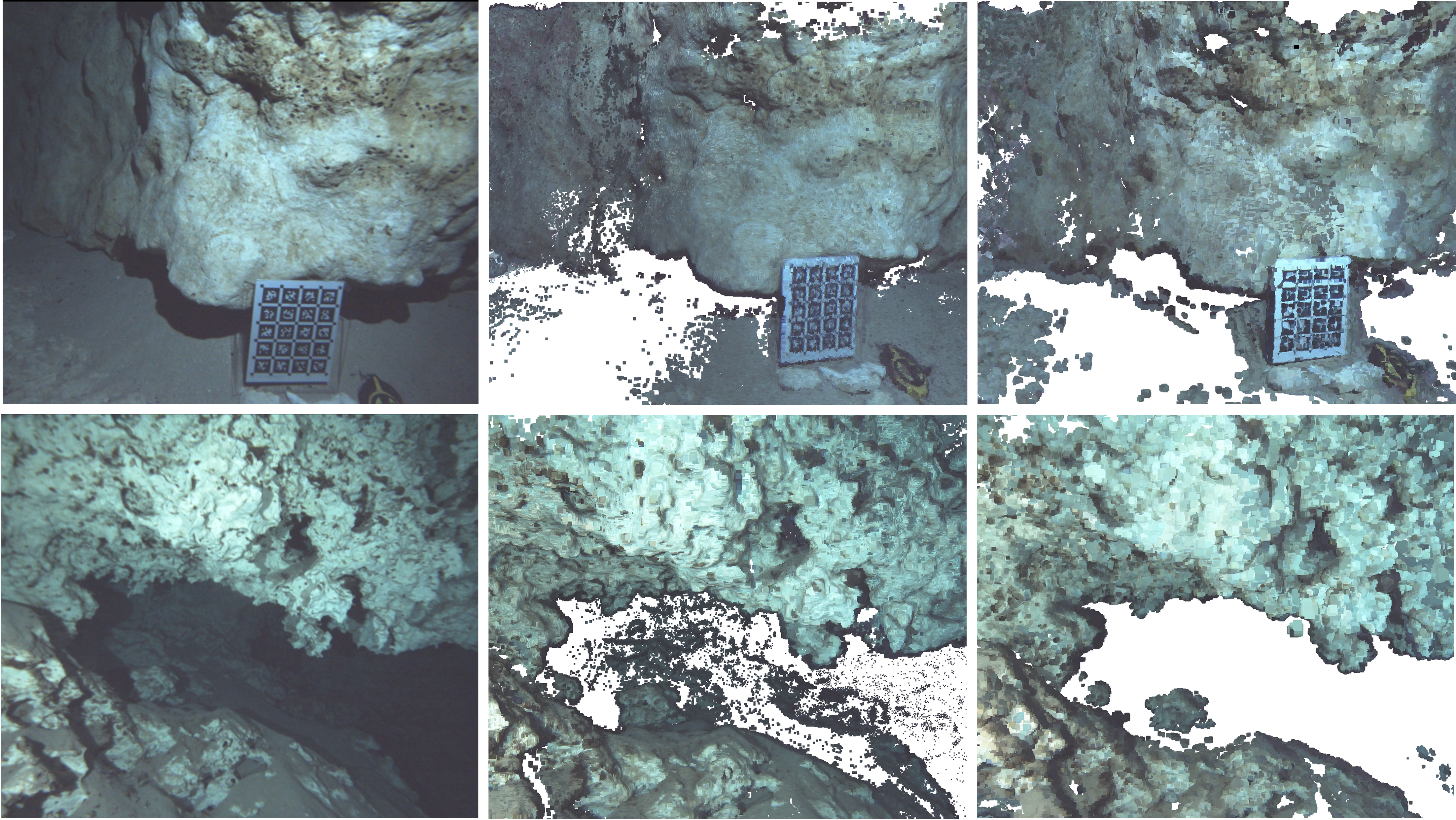}\\
         \hspace{0.3cm} Image \hspace{1.2cm} COLMAP\hspace{1.3cm} Pipeline
    \caption{Ginnie Ballroom (top), and Cenote (bottom).}
    \label{fig:ginnie_cenote_pc}
\end{figure}


\noindent \textbf{Reconstruction results}
Table \ref{tab:depth_map_results} summarizes the comparison of both the \textit{raw} and \textit{fused} depth maps from the pipeline with the geometric depth maps generated by COLMAP. The fused depth maps produced from COLMAP and the pipeline are similar for Gennie Ballroom and Cenote datasets with \textit{median-of-medians} 
and MAE in the 0.06 - 0.14 m range. The depth maps for the Coral Reef and Stavronikita datasets differ more, with both metrics in the 0.3 - 0.4 m range. \fig{fig:ginnie_depth} and \fig{fig:cenote_depth} show qualitative results, including raw and fused depth maps from the pipeline and depth maps from COLMAP. The COLMAP depth maps, while dense, contain noisy artifacts in open regions of the scene, typically resulting from floating particles and changes in illumination. The raw depth maps generated by the pipeline are also noisy, but the fusion step removes noise and produces depths corresponding to surfaces in the environment. 

\begin{table}[b]
    \centering
    \caption{Point cloud evaluation between COLMAP and Pipeline.}
    \label{tab:point_cloud_result}
    \resizebox{\columnwidth}{!}{
    \begin{tabular}{@{}l|cc|cc}
    \specialrule{.8pt}{0pt}{0pt}
    \multirow{2}{*}{\textbf{Dataset}} &  \multicolumn{2}{c|}{\textbf{pipeline-to-colmap}} & \multicolumn{2}{c}{\textbf{colmap-to-pipeline}} \\
     & Precision (\%) & Accuracy (m) & Recall (\%) & Completeness (m) \\
    \hline
     Ginnie Ballroom & 96.9 & 0.029 & 97.5 & 0.019\\
     Cenote & 94.4 & 0.037 & 92.4 & 0.047\\
     Coral Reef & 52.3 & 0.109 & 60.3 & 0.114\\
     Stavronikita & 43.6 & 0.134 & 40.2 & 0.143 \\
    \specialrule{.8pt}{0pt}{2pt}
    \end{tabular}
    }
\end{table}

\begin{figure}
\footnotesize
    \centering
        \includegraphics[width=0.95\columnwidth]{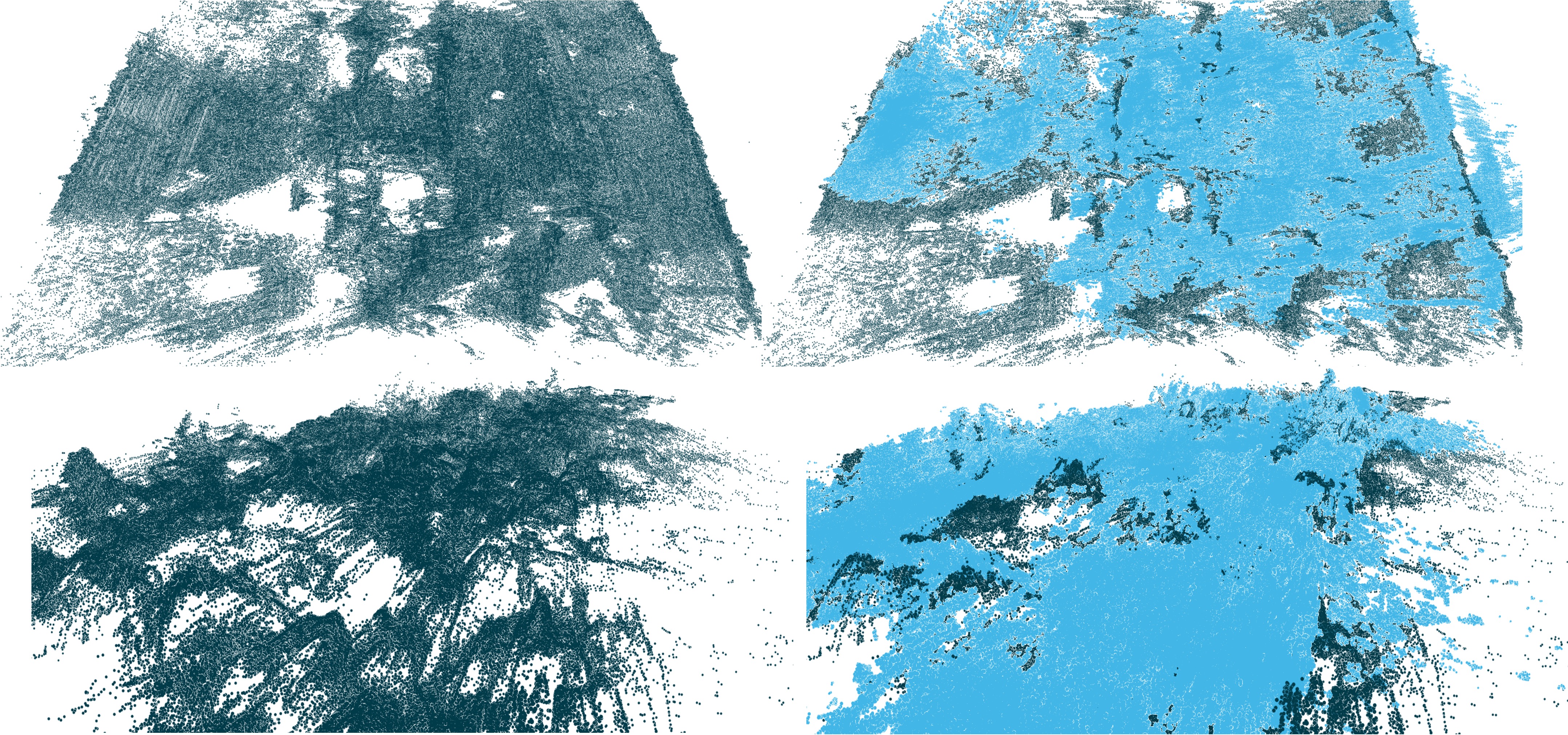} \\
        \hspace{0.2cm} COLMAP \hspace{2.5cm} COLMAP+Pipeline
    \caption{\textbf{Stavronikita Shipwreck} (top). \textbf{Coral Reef} (bottom). COLMAP reconstruction results overlayed with the Pipeline reconstruction. }
    \label{fig:stav_pc}
\end{figure}

\begin{figure}[t]
\footnotesize
    \centering
    \begin{tabular}{cc}
        \includegraphics[width=0.4\columnwidth]{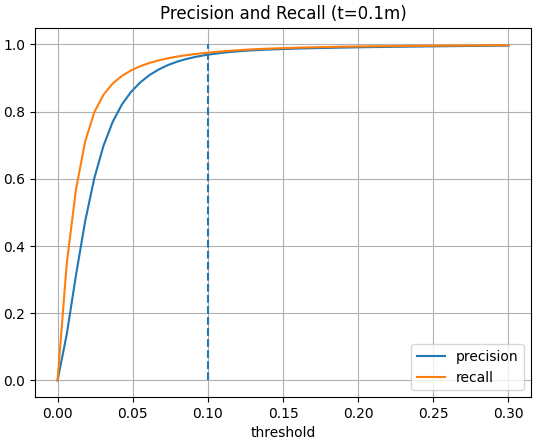}&
        \includegraphics[width=0.4\columnwidth]{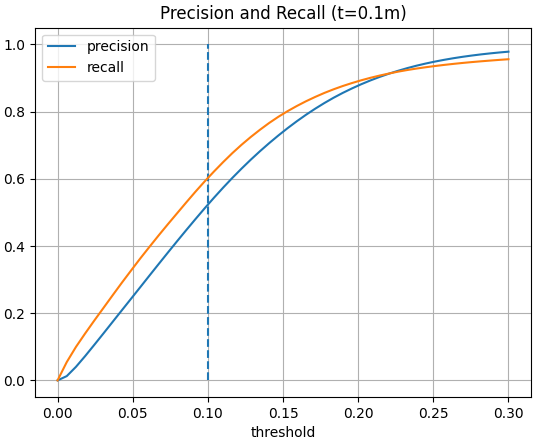}\\
         (a) Ginnie Ballroom & (b) Coral reef
    \end{tabular}
    \caption{Precision and Recall plots for the Ginnie Ballroom and Coral Reef datasets.}
    \label{fig:prec_rec_curves}
\end{figure}

In Table \ref{tab:point_cloud_result}, we show a quantitative comparison between pipeline and COLMAP point clouds. For the Ginnie Ballroom and Cenote datasets, the pipeline generates models that are very close to COLMAP with both accuracy and completeness less than 0.05 m, and precision and recall over 90\%. \fig{fig:ginnie_cenote_pc} shows a local perspective of the generated point clouds from the Ginnie Ballroom and Cenote datasets. Much of the detail is preserved in the pipeline reconstruction, with sparsity in areas of open water and smooth surfaces. The reconstruction results for Stavronikita and Coral Reef are 
inferior with accuracy and completeness less than 0.15 m, and precision and recall in the range 40\%-60\%. This can be explained in part by the sparsity of our models for these two datasets. The point clouds of the Stavronikita and Coral Reef datasets can be seen in \fig{fig:stav_pc} where the pipeline model is overlayed on the COLMAP point cloud. \fig{fig:prec_rec_curves} shows examples of the Precision and Recall curves for the Ginnie Ballroom and the Coral Reef as a function of the threshold. (The default 0.1 m is marked with vertical lines.)

It should be noted that the Ginnie Ballroom and Cenote datasets are collected by a slowly moving human diver with artificial illumination. The Coral Reef and Stavronikita datasets are collected by a fast-moving Aqua2 AUV in deep ocean without any artificial lightning and contain irregular surfaces. Thus, the images in the latter two datasets suffer from motion blur and color saturation. 
This leads to noisy point clouds by both systems and larger discrepancies between them. COLMAP is somewhat more robust, but its models are far from perfect on these data.

\noindent \textbf{Real-Time Reconstruction using SVIn2 Poses}
In the last experiment, we use poses obtained by SVIn2 as input to the pipeline to simulate actual deployment of our approach.
(A comparison of pose estimation results between COLMAP and SVIn2 is presented in Section \ref{sec:tracking}.) To account for discrepancy in dense reconstruction resulting from camera pose tracking error, we use the RMSE error between SVIn2 and COLMAP poses as a threshold to compute precision and recall as shown in Table \ref{tab:point_cloud_result_svin}. The results show that the dense reconstructions obtained using SVIn2 poses are accurate compared to those of COLMAP, with both precision and recall over 80\% for all datasets. The pipeline results show that even with drift in SVIn2 poses we are able to produce comparable reconstruction to that of COLMAP. This paves the way for real-time reconstruction 
onboard an Aqua2 AUV. 

\begin{table}[h]
    \centering
    \caption{Point cloud comparison between COLMAP and Pipeline. Both systems use SVIn2's poses without bundle adjustment.}
    \label{tab:point_cloud_result_svin}
    \resizebox{\columnwidth}{!}{
    \begin{tabular}{@{}l|c|c|c}
    \specialrule{.8pt}{0pt}{0pt}
    \multirow{2}{*}{\textbf{Dataset}} & \textbf{threshold} &  \textbf{pipeline-to-colmap} & \textbf{colmap-to-pipeline} \\
     & (m) & Precision (\%)  & Recall (\%)  \\
    \hline
     Ginnie Ballroom & 0.07 & 85.6 & 92.0 \\
     Cenote & 0.19 & 91.2 & 89.0\\
     Coral Reef & 0.39 & 81.8 & 85.9\\
    \specialrule{.8pt}{0pt}{0.1pt}
    \end{tabular}
    }
\end{table}

\vspace{-0.2in}
\section{CONCLUSIONS}

We have shown on a variety of challenging datasets that an online 
3D reconstruction system  with  robust VIO~\cite{RahmanIJRR2022} can obtain results on par with a much slower offline system. Such an evaluation was missing from the literature and helps answering the question on whether real-time dense reconstruction is feasible onboard. 
Dense 3D representations of the environment estimated in real time will enable improved navigation~\cite{XanthidisICRA2020} and autonomous operations for the Aqua2 AUV~\cite{SattarIROS2009}. Furthermore, gaps and boundaries of the dense reconstruction will effectively guide the AUV towards frontier points~\cite{XanthidisIROS2021} to enable mapping of underwater structures~\cite{XanthidisISRR2022}.

\newpage
\printbibliography

\end{document}